# Abnormal Spatial-Temporal Pattern Analysis for Niagara Frontier Border Wait Times

**Zhenhua Zhang[1], Lei Lin[2*]**
1. Civil, Structural & Environmental Engineering, University at Buffalo
2*. llin22@buffalo.edu; 585-489-2347;
Conduent Labs-US, Building 128, 800 Phillips Road, Webster, NY, 14580

**Abstract**
Border crossing delays cause problems like huge economics loss and heavy environmental pollutions. To understand more about the nature of border crossing delay, this study applies a dictionary-based compression algorithm to process the historical Niagara Frontier border wait times data. It can identify the abnormal spatial-temporal patterns for both passenger vehicles and trucks at three bridges connecting US and Canada. Furthermore it provides a quantitate anomaly score to rank the wait times patterns across the three bridges for each vehicle type and each direction. By analysing the top three most abnormal patterns, we find that there are at least two factors contributing the anomaly of the patterns. The weekends and holidays may cause unusual heave congestions at the three bridges at the same time, and the freight transportation demand may be uneven from Canada to USA at Peace Bridge and Lewiston-Queenston Bridge, which may lead to a high anomaly score. By calculating the frequency of the top 5% abnormal patterns by hour of the day, the results show that for cars from US to Canada, the frequency of abnormal waiting time patterns is the highest during noon while for trucks in the same direction, it is the highest during the afternoon peak hours. For Canada to US direction, the frequency of abnormal border wait time patterns for both cars and trucks reaches to the peak during the afternoon. The analysis of abnormal spatial-temporal wait times patterns is promising to improve the border crossing management.

## Abnormal Spatial-Temporal Pattern Analysis for Niagara Frontier Border Wait Times

### Introduction

Western New York has recently been recognized as being a part of what is now called the "Golden Horseshoe", a densely populated and industrialized region which encompasses Ontario and parts of New York State including the Buffalo‑Niagara Region. The economic vitality of the "Golden Horseshoe" is heavily dependent upon the ability to move goods freely and efficiently across the Canadian‑US border. This highlights the critical importance of the Niagara Frontier border, one of the busiest international crossings in the world. A report by the Ontario Chamber of Commerce (OCC) in 2005 puts the value of the annual land-borne merchandise crossing the Niagara Frontier border at $60.3 billion dollars *(1)*.

The Niagara Frontier border crossings include three main bridges connecting Western New York to Southern Ontario namely the Lewiston-Queenston Bridge (LQ), the Rainbow Bridge (RB), and the Peace Bridge (PB) as shown in Figure 1.

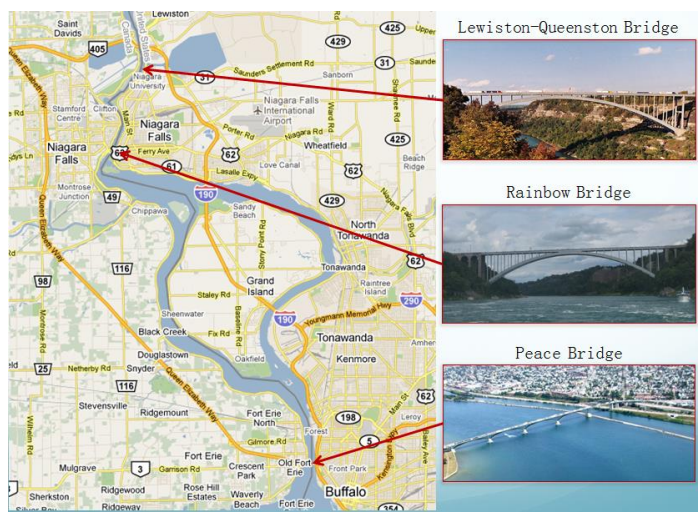

**Figure 1 the Locations of the Three Bridges**

However, due to the continuous travel demand increase, coupled with tighter security and inspection procedures after September 11, border crossing delay has become a critical problem. As reported by the Ontario Chamber of Commerce, border crossing delay causes an annual loss of approximately $268.45 million for New York State. For the whole U.S., the cost is much higher *(1)*. According to a press release in 2008 given by the then U.S. Transportation Secretary, Mary E. Peters, the US-bound traffic from Canada encountered delays as high as three hours at several crossings, with delays costing businesses on both Canadian and the US sides as many as 14 billion dollars in 2007 *(2)*.

To address these issues, transportation authorities have recently begun to provide travellers with information about current border crossing delays. This is the case for example in the Buffalo-Niagara region, for example, where the Niagara International Transportation Technology Coalition (NITTEC), a coalition of fourteen different agencies in Western New York and Southern Ontario, has been providing such information to the public for years. In the early years, the waiting time was obtained based on very rough and approximate estimates of queue length. More recently, NITTEC is using blue-tooth identification technology to provide more accurate delay estimates to motorists, and the information is now updated every five minutes.

Previously, the authors of this paper have proposed a two-step delay prediction model that consists of a short-term traffic volume prediction model for predicting the incoming traffic flow and a queueing model for predicting border inspection resulted delays *(3-7)*. An Android smartphone application called the Toronto Buffalo Border Wait Time (TBBW) app is also developed to collect, share and predict waiting time at the three Niagara Frontier border crossings *(8)*.

Recently a database has been built to store the wait time data collected and updated every five minutes by the Niagara Frontier border crossing authorities. This paper aims to identify and analyse the abnormal spatial-temporal wait time patterns across the three bridges for both passenger vehicles and commercial trucks in two directions. To find and compare the traffic delay patterns in three bridges, we employed a compression-based method to process the large volume of traffic delay data and extract the useful pattern information. The compression-based approaches are widely acknowledged in pattern recognition and anomaly detection. They have been proved valid in the field of system query processing *(9)*, signal transmission *(10)*, etc. The main idea of compression is to identify the frequent patterns and extract abnormal patterns from the data, which aims at finding the best pattern table to represent the original data. This is quite different from our previous studies, for which only the historical traffic volume data are utilized and the two-step delay prediction models are only built for the passenger vehicles from Canada to US at Peace Bridge.

The paper is organized as below. Section Two introduces the methodology to identify abnormal spatial-temporal patterns. Section Three describes the border wait time data in detail. Model results and insights discerned from these abnormal spatial-temporal patterns are discussed in Section Four. Finally, the paper ends with conclusions and suggestions for future work.

## Methodology

This section briefly discusses the procedure of employing the compression methods to quantify the wait time patterns on three Niagara Frontier border crossing bridges and detect the pattern anomalies from the border wait times data.

The first step is to discretize the continuous border wait times. Based on our previous studies, the wait times are categorized according to the following rules in Table 1:

**Table 1 Discretization of Border Wait Times**

| Category Index | Category Name | Wait Time $t$ (minutes) |
|---|---|---|
| 1 | no waiting | $t = 0$ |
| 2 | slight delay | $t > 0 \; and \; t \leq 15$ |
| 3 | delay | $t > 15 \; and \; t \leq 30$ |
| 4 | heavy delay | $t > 30$ |

The next step after categorization is to build a Database Table (DT) and Pattern Table (PT). Table 2 shows an example of DT and its PT and also illustrates one possible way to compress the DT by PT. First, in DT, each row of $G_i$, $g_i$, $\cancel{g}i$, $i = 1,2,3,4$ in the first column represents the categorized wait times from the three bridges PB, LQ and RB at the same time point. The second column of DT shows the corresponding PT pattern for each row. Second, the first column in PT is the PT pattern while the second column is the total usage of that pattern. Take the first four rows in DT as an example, all of them have the same combination of {$G_1$, $g_1$, $\cancel{g}1$} and can be represented by the same pattern $PF_1$. In PT, we can check the meaning of $PF_1$, and the total usage is 4.

**Table 2 an Illustrative Example of Database Table and Pattern Table**

| Database Table (DT) | | | | Pattern Table (PT) | |
|---|---|---|---|---|---|
| DB pattern ($DF_i$) | | | PT pattern included | PT pattern ($PF_i$) | usage of PT pattern |
| PB | LQ | RB | | | |
| $G_1$ | $g_2$ | $\cancel{g}1$ | $PF_1$ | $PF_1$: G1, g2, $\cancel{g}1$ | 4 |
| $G_1$ | $g_2$ | $\cancel{g}1$ | $PF_1$ | $PF_2$: G1, $\cancel{g}2$ | 2 |

| $G_1$ | $g_2$ | g̸1 | $PF_1$ | $PF_3$: $g_2$ | 2 |
|---|---|---|---|---|---|
| $G_1$ | $g_2$ | g̸1 | $PF_1$ | | |
| $G_1$ | $g_2$ | g̸2 | $PF_2, PF_3$ | | |
| $G_1$ | $g_2$ | g̸2 | $PF_2, PF_3$ | | |

Therefore the PT performs as a code dictionary and the process of converting the DB patterns into combination of PT patterns is called dictionary-based compression. There are 4 important indexes in describing the compression method:

1) The length of PT pattern ($PF_i$) in a certain pattern table is defined as:

$$L(usage\ (PF_i)|PT) = -log\left(\frac{usage\ (PF_i)}{\sum_{PF \in PT} usage\ (PF)}\right) \tag{1}$$

2) The length of DT pattern ($DF_i$) is calculated as the sum of the lengths of all PT patterns it contains:

$$L(DF_i|PT) = \sum_{PF_j \in DF_i} L\big(usage\ (PF_j)|PT\big) \tag{2}$$

3) The length of DT is the sum of the lengths of all DT patterns :

$$L(DT|PT) = \sum_{DF_i \in DT} L(DF_i|PT) \tag{3}$$

4) The length of the PT is defined as:

$$L(PT) = \sum_{PF_i \in PT} L(usage\ (PF_i)|PT) + \sum_{r_i \in I} -r_i \log\left(\frac{r_i}{c}\right) \tag{4}$$

There are two parts involved in the length of PT. The first part in Equation (4) is the sum of lengths of all PT patterns, while the second part is the sum of lengths of all singleton items in each category in DB. $I$ is defined as all the singleton items in DT, $c$ is the total count of singleton items and $r_i$ is the count of the ith singleton item. For example, in Table 2, $c$ is equal to 18, $r_i$ of the singleton item "$G_1$" is 6. One can see that PT plays a vital role in describing the DT.

The compression technique seeks the best length of codes to represent the DB patterns. The previous large data observations can be then converted into a relatively small one with all details. A relatively more detailed PT can compress the DB better but results into a relatively larger PT while a less detailed PT will just do the reverse. Here, we need to use the Minimum Description Length (MDL) to find a balanced PT. The principle can be written as:

$$min\ \mathcal{L} = L(DT|PT) + L(PT) \tag{5}$$

Where $\mathcal{L}$ is the total length that should be minimized. We employed the algorithm proposed by *(11)* to find the best set of PT, the pseudo code of which is shown in Table 3. The algorithm employs the Apriori Algorithm *(12)* to find the best combination of the DT patterns.

**Table 3 Pseudo Code of Dictionary-based Compression**

**Input**: Database with n rows and m categories
**Output**: A PT table and the usage of each pattern

**Build** the initial PT table and all PT patterns $PF_i$ are singleton items of features in DB
**Compute** the initial description length $\mathcal{L}_0$, the optimal length $\mathcal{L} = \mathcal{L}_0$
**Implement** the Apriori algorithm to find all frequent items $FI$ whose frequency is higher than a threshold $T$, these frequent items constitute a set $S$

```
Repeat
    for FI_i in S
        Put FI_i into the PT table
        Compute the current description length ℒ_i
        If ℒ_i < ℒ
            ℒ = ℒ_i
            remove FI_i from S
            add FI_i into PT table
        else
            remove FI_i from S
until |S| = 0
```

The algorithm can find the best PT which minimizes the MDL based on wait times data of passenger vehicles or trucks from the same direction at three bridges. The MDL algorithm calculates the best PT table that quantify a set of $DF_i$ and can extract the abnormal $DF_i$. These abnormal $DF_i$ should be the irregularly feature combination on three bridges. For instance, during certain time-of-day, there is a surge of traffic in a certain day as compared to moderate traffic in other days. The traffic pattern on that day should be taken as abnormal patterns

Our algorithm can find the abnormal patterns. According to Equation (2), the length of DT pattern $(DF_i)$ is calculated based on the frequencies of the $PF_j$ that it contains. Commonly-seen patterns should have a short length while less frequent ones should do the opposite. MDL criteria makes the best estimation of each pattern length. They are expected to work well in extracting the abnormal traffic patterns in three border-crossing bridges.

## Data Description

In the current study, the dataset is comprised of wait times for two vehicle types (passenger vehicle and commercial vehicle), two directions (to US and to Canada), and three bridges from 08/22/2016 to 01/16/2017. For the LQ and PB, the wait times are updated every five minutes, in total 40,243 observations are recorded. For RB, the wait times are updated hourly because the Bluetooth technology is not available, except that, the trucks are not permitted to go through RB. To eliminate the effects of fluctuation and match the wait times of the three bridges, we replace the five-minute wait times from LQ and PB with hourly wait times by taking the average for each hour.

Some empirical results of the delay characteristics on three bridges are presented. Figure 2(a) and 2(b) shows the delay distribution from US to Canada, separately for cars and trucks. One can see that under most conditions, both the cars and trucks have no delay. If we define the border crossing pattern as the combination of delay status on three bridges, it can be expected that both the patterns and the delay status on three bridges are different during different time periods. Those patterns that are extremely different from the rest of the other time periods should be taken seriously and the corresponding features should be unveiled.

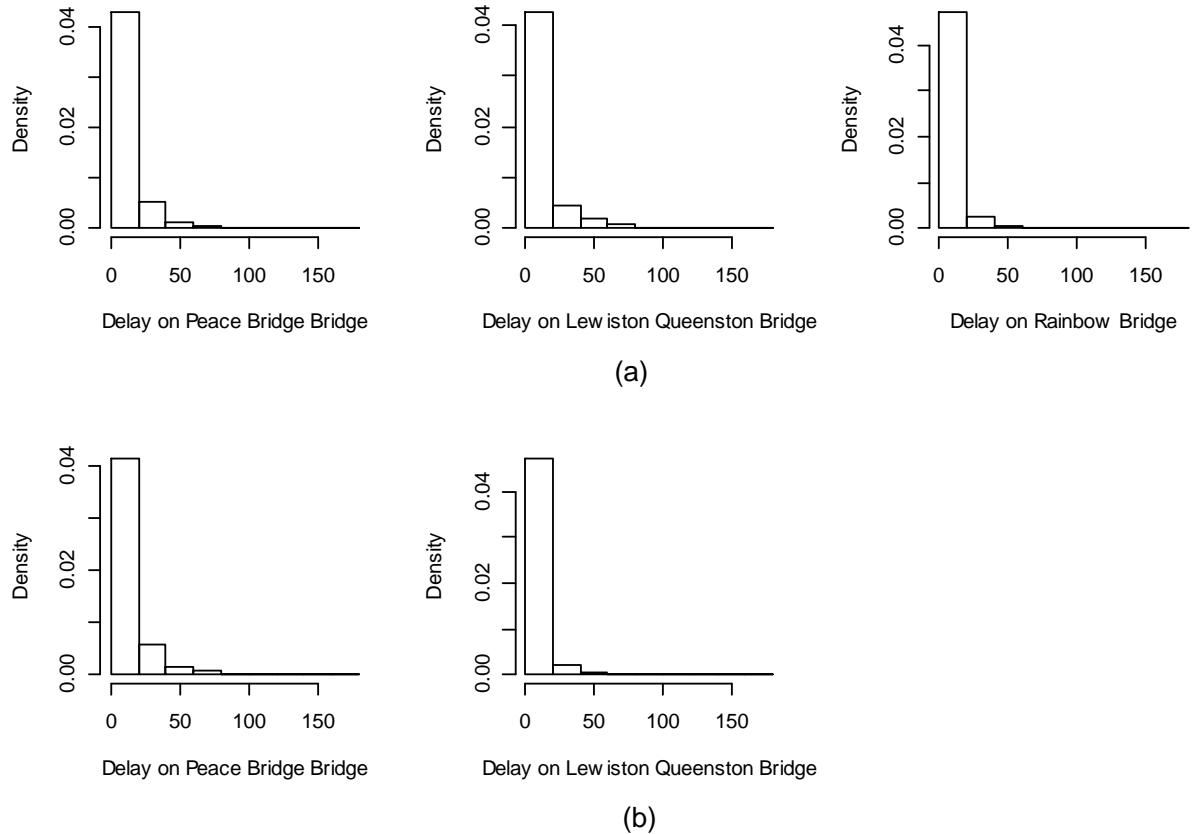

(a)

(b)

**Figure 2 Delay Distribution of Three Bridges from USA to Canada for (a) cars and (b) trucks.**

As another example of empirical analysis, the following Figure 3 shows the categorized delay type distribution by hour of the day (7:00 -21:00) for passenger vehicles and trucks from two directions at Peace Bridge. As can be seen, the wait time patterns vary with the vehicle type and the direction. For example, comparing Figure 2(a) with 2(b), the passenger vehicles to Canada only have a percentage of around 40% to cross the border without waiting at 10:00 and 11:00, while it looks like the congestion happens at 18:00 and 19:00 for the passenger vehicles to USA and the percentages of higher waiting times are much lower. Furthermore, comparing Figure 2(a) with 2(c), for the same direction from US to Canada, the passenger vehicles mainly experience higher delays in the morning, but the rush-hour period is 13:00 – 17:00 for the trucks.

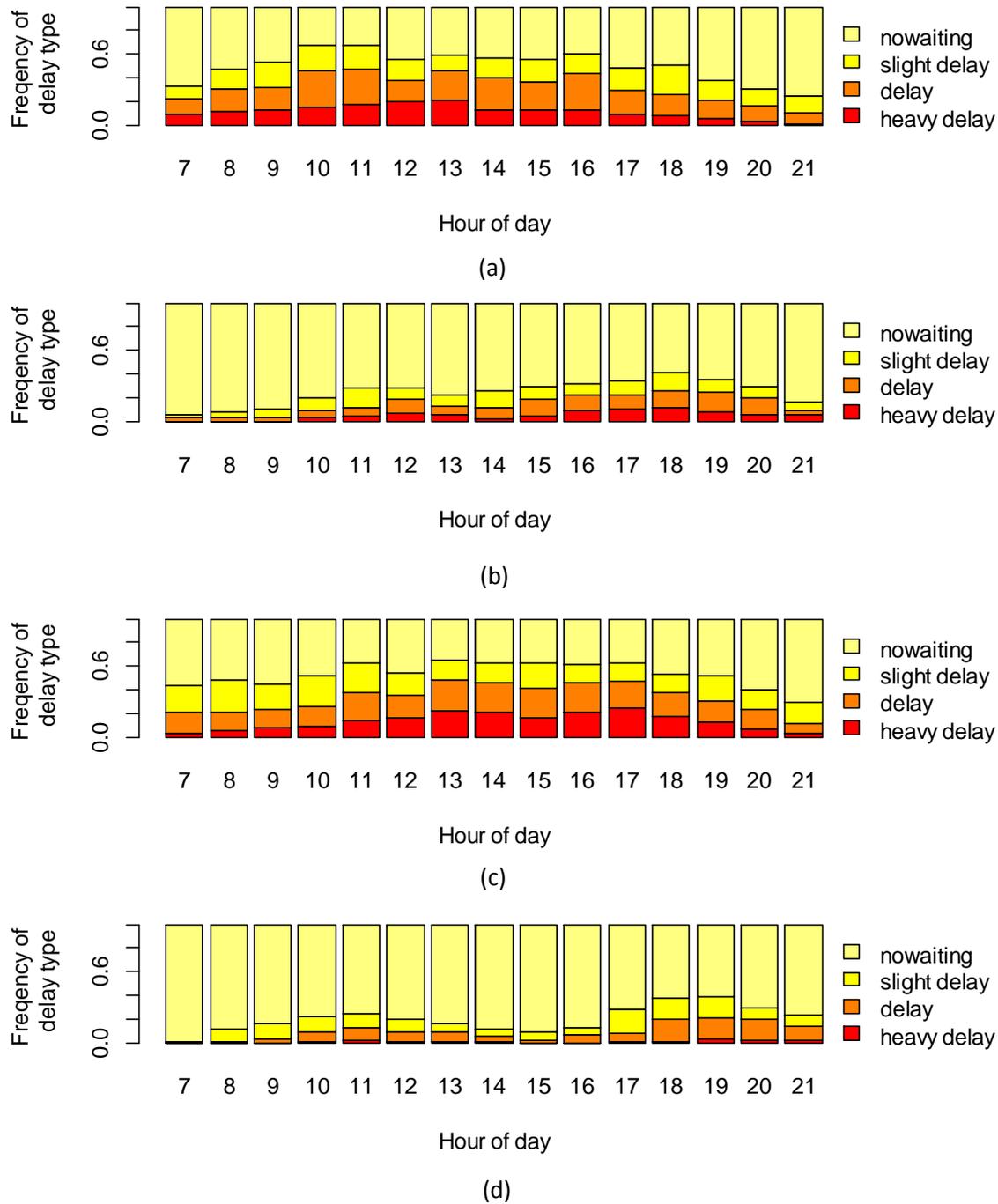

**Figure 3 Delay Type Distribution during Different Hours of Day at PB (a) cars from USA to Canada; (b) cars from Canada to USA; (c) trucks from USA to Canada; (d) trucks from Canada to USA.**

## Abnormal Pattern Analysis

As introduced in Methodology section, all the wait times are converted to category variables. The dictionary-based compression algorithm can further generate a DT table and PT table for wait times of each vehicle type and each direction at three bridges. All the border wait time patterns $DF_i$ during our observation periods are ranked according to their anomaly scores.

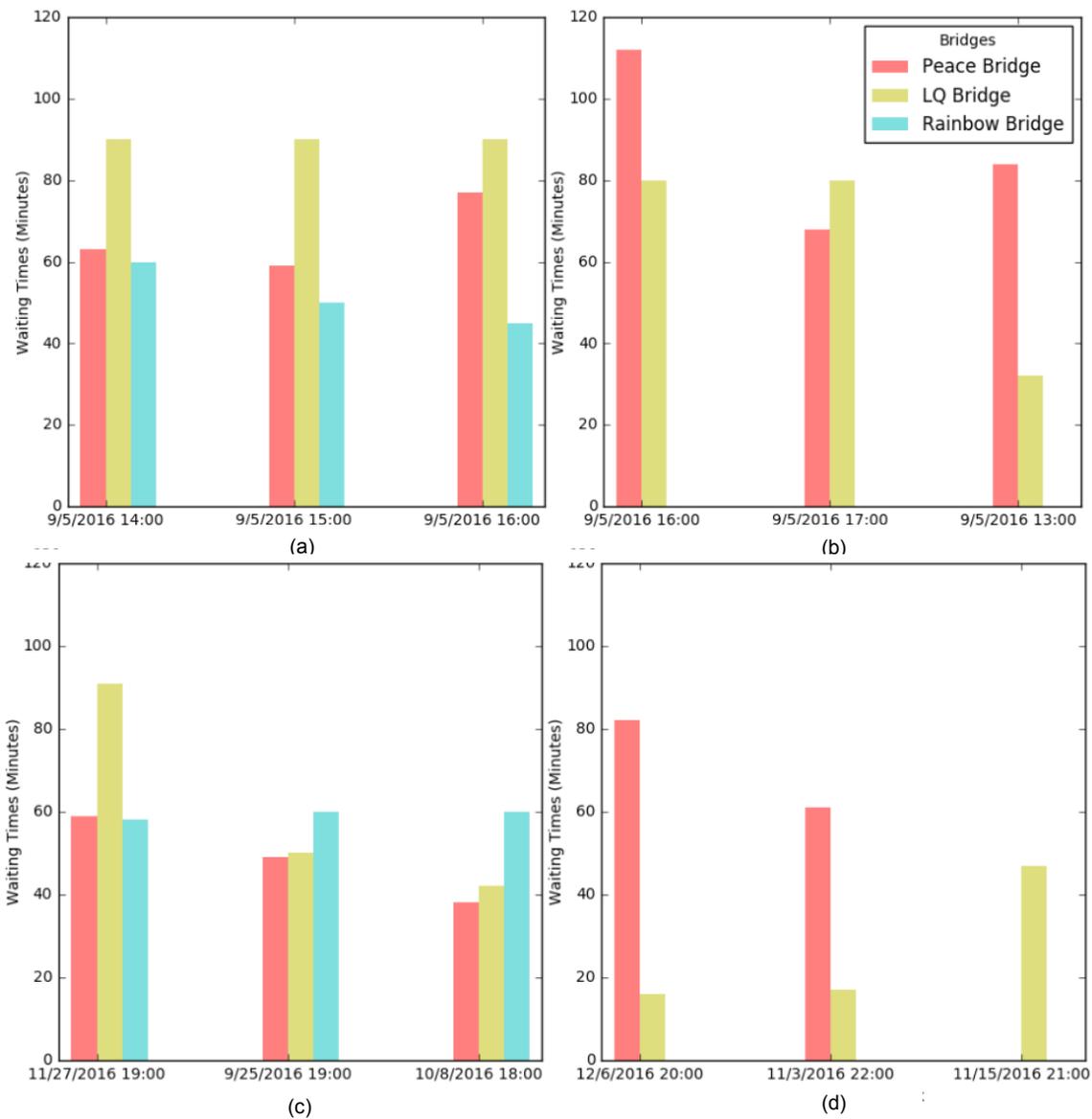

**Figure 4 Top Three Abnormal Spatial-Temporal Patterns for Different Scenarios (a) cars from USA to Canada; (b) trucks from USA to Canada; (c) cars from Canada to USA; (d) trucks from Canada to USA;**

Figure 4 shows the top three abnormal spatial-temporal patterns at three bridges for different scenarios. The date and time of these patterns are also labeled. Therefore some hidden reasons can be uncovered. First, Figure 4(a) shows that the top three abnormal patterns for passenger vehicles from USA to Canada all happened in the afternoon of 09/05/2016, Labor Day. All three bridges were in heavy congestion, especially for the LQ Bridge, where the delay was around 90 minutes. Second, the trucks through Peace Bridge and LQ Bridge from USA to Canada also encountered longer waiting times on Labor Day in 2016, which can be verified by the top three abnormal patterns in Figure 4(b). Third, the top three abnormal patterns for cars from Canada to USA in Figure 4(c) are from 11/27/2016 (Sunday), 09/25/2016 (Sunday) and 10/08/2016 (Saturday) separately, which are all weekends. It once again shows that for the passenger vehicles, the three bridges are always in heavy congestion at the same time. Last, different with the previous three figures, abnormal patterns in Figure 4(d) show the waiting times for trucks from Canada to USA at PB and LQ Bridge could be very different. As can be seen, for 20:00 on 12/06/2016 and 22:00 on 11/03/2016, the trucks at PB experienced much higher delay comparing with LQ Bridge. For 21:00 on 11/15/2016, the PB had no delay while the waiting time at the LQ Bridge was about 40 minutes for trucks from Canada to USA.

We further explored the time-of-day information of the abnormal border wait time patterns at three bridges. The top 5% most abnormal patterns are extracted. The frequency of these abnormal patterns are calculated by time-of-day and are shown in Figure 5. One can see different frequency distributions from US to Canada than from Canada to US for both the cars and vehicles. We can also find that:

- From US to Canada direction, abnormal border wait time patterns usually happen in the daytime. The frequency of abnormal patterns for cars is the highest during the noon, while for trucks it happens during PM peak.

- From Canada to US direction, the frequencies of abnormal border wait time patterns for both cars and trucks are the highest in the afternoon. For cars other abnormal patterns mainly happen during the noon, while for trucks those happen during 10:00.

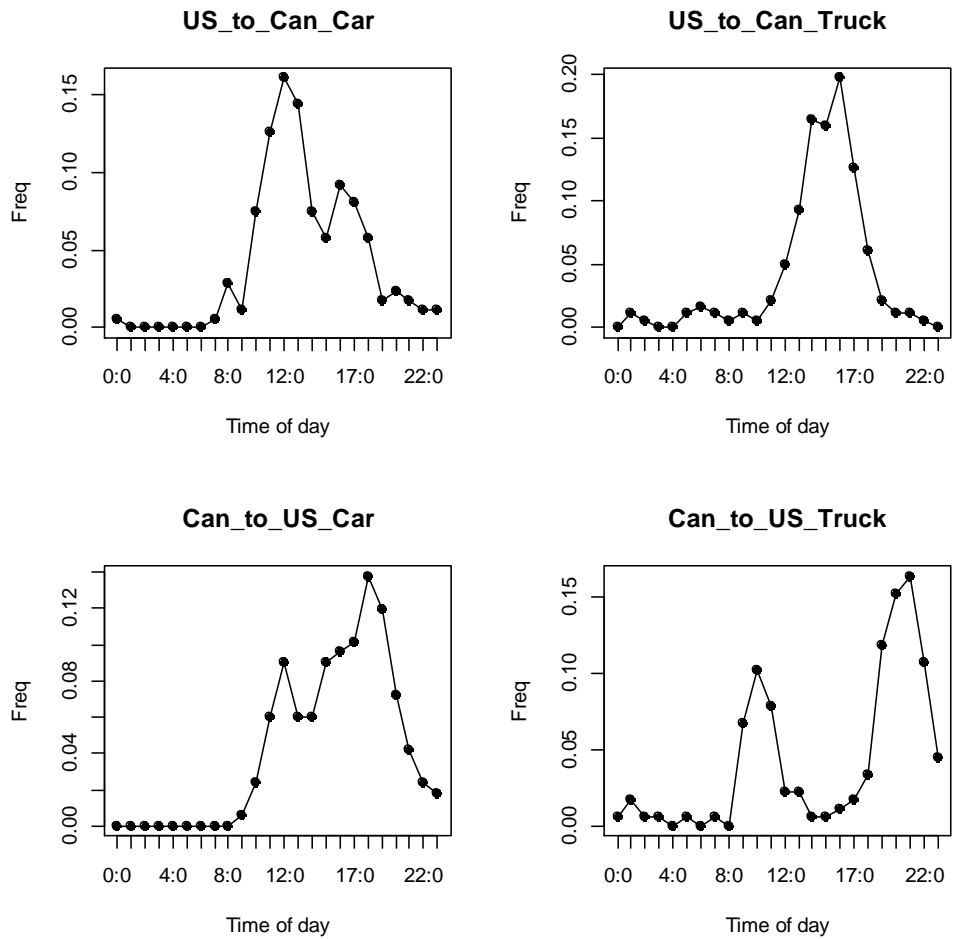

**Figure 5 Time-of-day Features of the Most Abnormal Traffic Patterns**

We can see that the ranking and analyzing of abnormal patterns uncover the weak points of the border crossing system, and therefore help the management authorities to make corresponding plans.

One can also see the advantages of our method is that it is effective in mining patterns and detecting anomalies which find useful knowledge from the large datasets by giving a good quantitative describing length. Another advantage is that both the compression method and MDL algorithm is easy to apply and the calculation is not time-consuming which is of great applicability in real-time application.

## Conclusion and Future Studies

In this paper, we employed a dictionary-based compression algorithm to process the spatial-temporal waiting time data at three Niagara Frontier border crossing bridges. We ranked and analyzed the abnormal patterns for both cars and trucks from two directions from US to Canada and from Canada to US. The dictionary-based compression algorithm can effectively capture the key hidden patterns, which is very meaningful in the big data era. Some main observations are summarized as following:

1. Based on the analysis of top three abnormal wait time patterns for different vehicles and directions, there are at least two reasons behind the anomaly. First the weekends and holidays may cause unusual heave congestions at the three bridges at the same time; second the uneven freight transportation demand from Canada to USA at PB and LQ may also lead to a high anomaly score.

2. By correlating the time-of-day information with the top 5% abnormal patterns, we find for US to Canada direction, the frequency of abnormal waiting time patterns is the highest during noon for cars while it is the highest during PM peak hours for trucks. For Canada to US direction, the frequency of abnormal border wait time patterns for both cars and trucks reaches to the peak during the afternoon.

For future studies, more features can be involved into the pattern identification and ranking including weather, traffic flow features, etc.

## References

1. Ontario Chamber of Commerce (OCC). (2005). Cost of Border Delays to the US Economy. Available online at:
http://www.thetbwg.org/downloads/Cost%20of%20Border%20Delays%20to%20the%20United%20States%20Economy%20-%20April%202005.pdf . Accessed on May 29, 2016.
2. U.S. Department of Transportation (USDOT). Office of Public Affairs. (2008). US Department of Transportation Unveils New Program to Fight Border Congestion. Available online at: http://www.dot.gov/affairs/fhwa1208.htm. Accessed on May 29, 2016.
3. Lin, L., A. W. Sadek, and Q. Wang (2012). Multiple-Model Combined Forecasting Method for Online Prediction of Border Crossing Traffic at Peace Bridge. *In Transportation Research Board 91st Annual Meeting*, No. 12-3398.
4. Lin, L., Q. Wang, & A. Sadek (2013). Short-term forecasting of traffic volume: evaluating models based on multiple data sets and data diagnosis measures. *Transportation Research Record: Journal of the Transportation Research Board*, (2392), 40-47.
5. Lin, L., Y. Li, and A. W. Sadek (2013). A k nearest neighbor based local linear wavelet neural network model for on-line short-term traffic volume prediction. *Procedia-Social and Behavioral Sciences*, 96, pp. 2066-2077.
6. Lin, L., Q. Wang, S. Huang, and A. W. Sadek (2014). On-line prediction of border crossing traffic using an enhanced Spinning Network method. *Transportation Research Part C: Emerging Technologies*, 43, pp. 158-173.
7. Lin, L., Q. Wang, & A. W. Sadek (2014). Border crossing delay prediction using transient multi-server queueing models. *Transportation Research Part A: Policy and Practice*, *64*, 65-91.
8. Lin, L., Q. Wang, A. W. Sadek, and G. Kott (2015). An Android Smartphone Application for Collecting, Sharing, and Predicting Border Crossing Wait Time. *In Transportation Research Board 94th Annual Meeting*, No. 15-1971.
9. Yan, H., S. Ding, T. Suel (2009). Inverted index compression and query processing with optimized document ordering, *Proceedings of the 18th international conference on World wide web*. ACM, pp. 401-410.

10. Lu, Z., D.Y. Kim, W.A. Pearlman (2000). Wavelet compression of ECG signals by the set partitioning in hierarchical trees algorithm. *IEEE transactions on Biomedical Engineering* 47(7), 849-856.
11. Zhang, Z., He, Q., Tong, H., Gou, J., Li, X. (2016). Spatial-temporal traffic flow pattern identification and anomaly detection with dictionary-based compression theory in a large-scale urban network. *Transportation Research Part C: Emerging Technologies* 71, 284-302.
12. Agrawal, R., Srikant, R., 1994. Fast algorithms for mining association rules, Proc. 20th int. conf. very large data bases, VLDB, pp. 487-499.